\documentclass[letterpaper, 10pt, journal, twoside]{IEEEtran}

\usepackage{amsmath}
\usepackage{graphicx} 
\newcommand{\blfootnote}[1]{%
  \begingroup
  \renewcommand\thefootnote{}
  \footnote{#1}%
  \addtocounter{footnote}{-1}%
  \endgroup
}
\usepackage[
    backend=biber,
    style=numeric-comp,
    sorting=none,
    maxbibnames=1,
    minbibnames=1
]{biblatex}

\bibliography{sample}
\usepackage{caption,subcaption}
\usepackage[colorlinks]{hyperref}
\hypersetup{
    colorlinks=true,
    urlcolor=black
}
\usepackage{booktabs}
\usepackage{multirow}
\DeclareMathOperator*{\argmax}{argmax} 
\title{\LARGE \bf
Enhancing Multi-Robot Exploration Using Probabilistic Frontier Prioritization with Dirichlet Process Gaussian Mixtures
}

\author{John Lewis Devassy$^{1}$, Meysam Basiri$^{1}$, M\'ario A. T. Figueiredo$^{2}$, and Pedro U. Lima$^{1}$
\thanks{Manuscript received: February 4, 2026; Revised: April 14, 2026; Accepted: June 1, 2026.}%
\thanks{This paper was recommended for publication by Editor M. Ani Hsieh upon evaluation of the Associate Editor and Reviewers comments.}%
\thanks{This work was supported in parts by a doctoral grant from Funda\c{c}\~ao para a Ci\^encia e a Tecnologia (FCT) by project reference UI/BD/153758/2022, in part by the Aero.Next project (PRR - C645727867-00000066), and in part by ISR/LARSyS and IT Strategic Funding through the FCT project (DOI: \url{https://doi.org/10.54499/UIDB/50009/2020}, \url{https://doi.org/10.54499/UIDP/50009/2020}, \url{https://doi.org/10.54499/LA/P/0083/2020}, and \url{https://doi.org/10.54499/UID/50008/2025}).}%
\thanks{$^{1}$J. L. Devassy, M. Basiri, and P. U. Lima are with the Institute for Systems and Robotics / LARSyS and Instituto Superior T\'ecnico, Universidade de Lisboa, Lisbon 1049-001, Portugal
(e-mail: john.lewis@tecnico.ulisboa.pt, meysam.basiri@tecnico.ulisboa.pt, and pedro.lima@tecnico.ulisboa.pt).}%
\thanks{$^{2}$M. A. T. Figueiredo is with the Instituto de Telecomunica\c{c}\~oes and Instituto Superior T\'ecnico, Universidade de Lisboa, Lisbon 1049-001, Portugal
(e-mail: mario.figueiredo@tecnico.ulisboa.pt).}%
\thanks{Digital Object Identifier (DOI): see top of this page.}%
}

\begin{document}

\maketitle

\begin{abstract}
Multi-agent autonomous exploration is essential for applications such as environmental monitoring, search and rescue, and industrial-scale surveillance. However, effective coordination under communication constraints remains a significant challenge. Frontier exploration algorithms analyze the boundary between the known and unknown regions to determine the next-best view that maximizes exploratory gain. This article proposes an enhancement to existing frontier-based exploration algorithms by introducing a probabilistic approach to frontier prioritization. By leveraging Dirichlet process Gaussian mixture model (DP-GMM) and a probabilistic formulation of information gain, the method improves the quality of frontier prioritization. The proposed enhancement, integrated into two state-of-the-art multi-agent exploration algorithms, consistently improves performance across environments of varying clutter, communication constraints, and team sizes. {Simulations showcase an average exploration time improvement of $10\%$ and $14\%$ for the two algorithms across all combinations.} Successful deployment in real-world experiments with a dual-drone system further corroborates these findings. 
\end{abstract}
\section{INTRODUCTION}

\label{sec:introduction}
The use of autonomous robots in search-and-rescue missions, disaster response, large-scale environment monitoring, and field inspections necessitates efficient exploration strategies. Exploration, a fundamental aspect of robotics, presents a complex and multi-faceted challenge, influenced by parameters such as speed, operational safety, robot capabilities, and environment complexity. In the context of large-scale outdoor exploration, a time-efficient solution opts for a coordinated multi-robot system rather than a single-robot alternative. {However, multi-robot exploration under communication constraints introduces additional complexities, such as delays or blackouts that stall information flow. In these scenarios, rigid task allocation often fails, causing robots to execute redundant, overlapping trajectories. By utilizing a probabilistic approach to task delegation, the system establishes continuous task delegation that prevents redundant routing.} Given these constraints, a fast multi-robot exploration algorithm that ensures robustness across varying environment features, while ensuring scalability and minimal parameter tuning, is essential.

Robotic exploration has been extensively studied in various contexts, including single-agent systems   \cite{stachniss2005information}, multi-agent systems \cite{lewis2024frontier}, indoor environments   \cite{burgard2005coordinated}, with a focus on enhancing exploration speed   \cite{zhou2023racer}, improving both speed and safety   \cite{bartolomei2023fast}, under communication-denied conditions   \cite{bartolomei2023fast,zhou2023racer}, and effective coordination   \cite{gao2021meeting}. While challenges are present in indoor scenarios  \cite{bigazzi2022focus}, outdoor exploration tasks raise additional constraints such as varying terrain types, large-scale areas with sparse or dense clutter, and limited communication infrastructure. Outdoor robotic tasks vary from environment monitoring   \cite{bartolomei2023fast} and field inspection  \cite{bettencourt2024geers} to disaster response and rescue missions \cite{ghassemi2022multi}. The environment of interest can be spread out, like in wind farm applications  \cite{khalid2022applications}, symmetric, such as in solar farms  \cite{bettencourt2024geers}, or varying clutter, like in forests/agriculture  \cite{ding2022recent}. Given such diversity, outdoor exploration tasks in robotics are challenging, with environment-specific algorithms  \cite{bartolomei2023fast} frequently employed to surpass the performance of generic approaches  \cite{zhou2023racer,burgard2005coordinated}.

\begin{figure*}[h]
    \centering
    \begin{subfigure}{0.30\textwidth}
        \centering
        \includegraphics[width=\textwidth,height=0.6\textwidth]{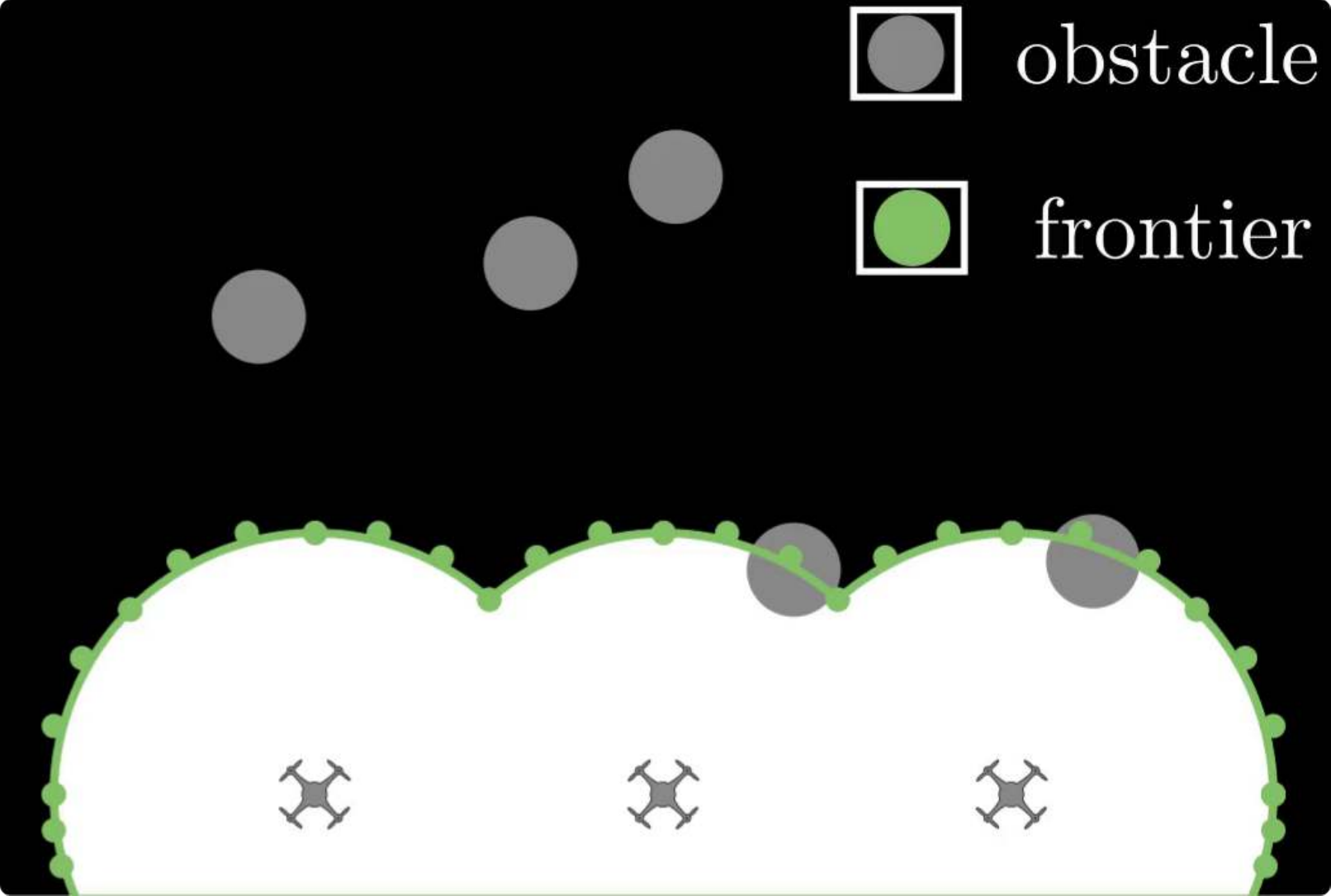}
        \caption{Frontier generation and discretization.}
        \label{fig:frontier_generation}
    \end{subfigure}
    \hfill
    \begin{subfigure}{0.30\textwidth}
        \centering
        \includegraphics[width=\textwidth,height=0.6\textwidth]{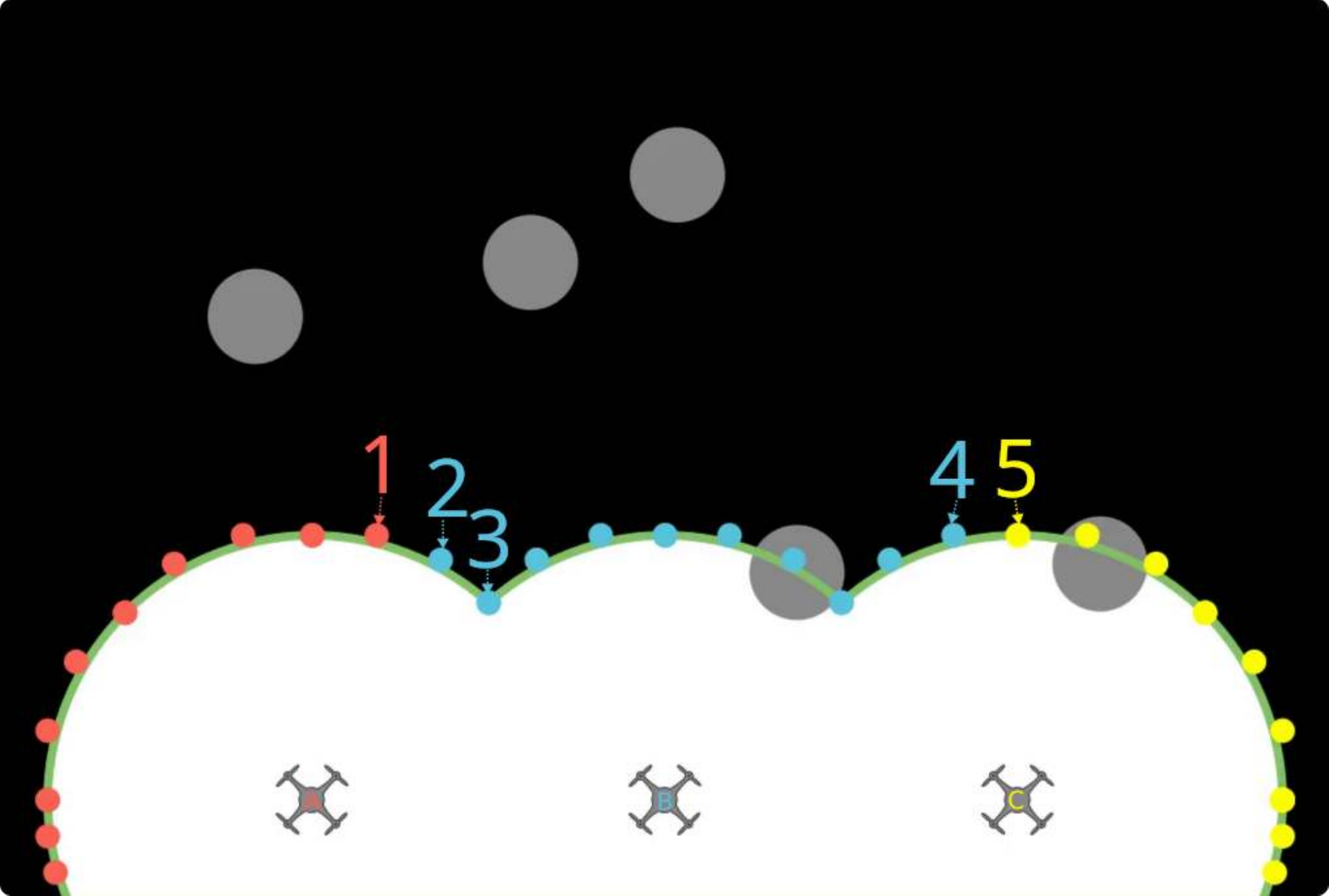}
        \caption{{Hard clustering with K-Means.}}
        \label{fig:frontier_clustering_knn}
    \end{subfigure}
    \hfill
    \begin{subfigure}{0.30\textwidth}
        \centering
        \includegraphics[width=\textwidth,height=0.6\textwidth]{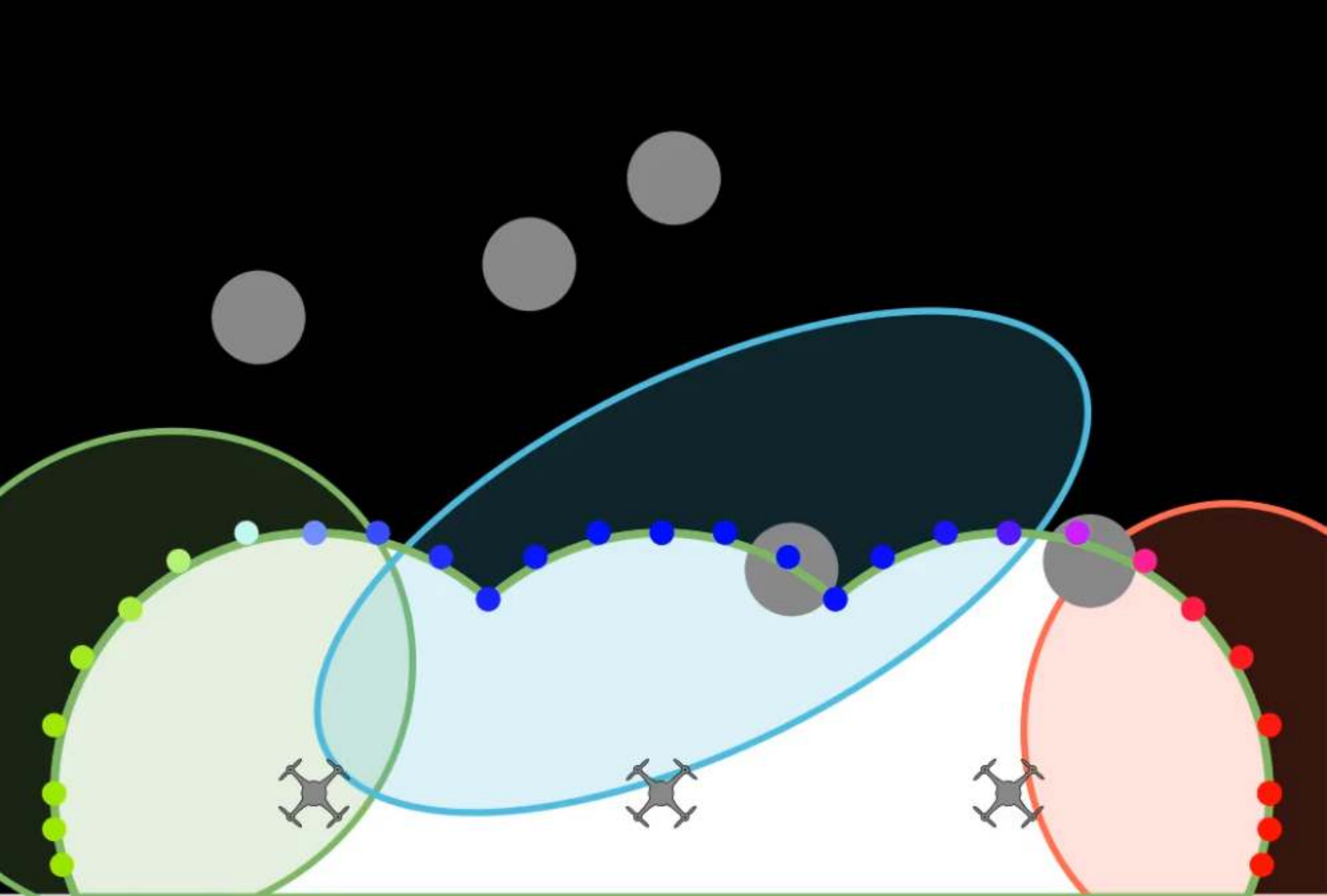}
        \caption{Soft clustering with GMM.}
        \label{fig:frontier_clustering_gmm}
    \end{subfigure}
\caption{(a) The white and black spaces represent the known and unknown, respectively. The continuous \textcolor{black}{green} boundary represents the frontier, and the \textcolor{black}{green} dots represent the discretized frontiers. (b) We can see that because of hard clustering, frontier \textcolor{black}{\textbf{1}} will be considered separate from frontier \textcolor{black}{\textbf{2}} and frontier \textcolor{black}{\textbf{3}}, which will lead to separate allocation after frontier processing. Similarly, despite frontier \textcolor{black}{\textbf{4}} being closer to drone \textcolor{black}{C} than drone \textcolor{black}{B}, the K-Means clustered it separately from frontier \textcolor{black}{\textbf{5}}. (c) The responsibility of \textcolor{black}{green}, \textcolor{black}{blue}, and \textcolor{black}{red} Gaussians in a given frontier is represented by the color saturation of the \textcolor{black}{green}, \textcolor{black}{blue}, and \textcolor{black}{red} respectively.}
    \label{fig:frontier_comparison}
\end{figure*}

A classical exploration approach identifies and evaluates frontiers to determine the next-best pose that maximizes exploratory gain  \cite{yamauchi1997frontier}. Frontiers are defined as the boundary between known and unknown regions of the map.
Frontiers are continuous in nature but are often represented as discrete cells in an occupancy grid   \cite{stachniss2005information}, or voxels in a pointcloud   \cite{bartolomei2023fast}. Fig. \ref{fig:frontier_generation} depicts the continuous and discrete frontiers for a three-robot team equipped with a $360^{\circ}$ perception sensor at the start of exploration. State-of-the-art large-scale exploration algorithms such as   \cite{bartolomei2023fast,zhou2023racer} use hard clustering to minimize the search space by creating viewpoints from frontiers. Viewpoints are defined as poses from which a cluster of frontiers can be covered, depending on the perception sensor's range and field of view. Distance-based hard clustering ensures well-defined exploration tasks; however, it can unintentionally assign nearby frontiers to different clusters or viewpoints, potentially leading to suboptimal exploration. This rigid partitioning may negatively impact multi-agent task allocation and exploration efficiency. Frontier clustering is carried out using either a non-parametric region-growing or a parametric K-means algorithm   \cite{bartolomei2023fast,zhou2023racer}. Fig. \ref{fig:frontier_clustering_knn} showcases hard K-means clustering of the frontiers depicted in Fig. \ref{fig:frontier_generation}. 

The hardness of K-means clustering can be overcome by adopting a soft clustering approach like Gaussian mixture models (GMM)   \cite{reynolds2009gaussian}. In robotics, GMMs have been employed in novelty detection in $3$-D maps   \cite{drews2013novelty,nunez2009novelty}, path planning   \cite{petrovic2022mixtures}, and localization   \cite{pfaff2008gaussian}. GMM's ability to represent large datasets compact, high-fidelity Gaussian distributions makes them particularly useful for information sharing in multi-robot systems  \cite{tabib2019real,wu2022mr}.
In this article, we use GMMs to improve frontier prioritization rather than as a mapping representation. In the context of frontier clustering, GMMs model the data as a mixture of Gaussian distributions. Rather than segregating the data points into specific clusters, a GMM determines the \textit{responsibility} of each Gaussian for a given data point. This probabilistic approach enables more flexibility in processing frontiers, as shown in Fig. \ref{fig:frontier_clustering_gmm}.

In an exploration objective, the frontier size varies as the robot navigates through the terrain, and a pre-constrained cluster count can lead to subpar task allocation. However, in its basic form, a GMM is a parametric clustering algorithm that requires knowledge of the number of clusters beforehand. To overcome this, we leverage the semi-parametric nature of Dirichlet process Gaussian mixture modeling (DP-GMM) to adapt the number of clusters to the data, avoiding the limitations of a fixed-complexity model.

\subsection{Contributions}
We propose a frontier prioritization technique for multi-robot exploration frameworks that:
\begin{itemize}
\item Utilizes a Dirichlet Process Gaussian Mixture Model (DP-GMM) to generate smooth cluster probabilities for task allocation;
\item Integrates a viewpoint processor that fuses entropy-based information gain with DP-GMM probabilities to generate improved frontier priorities;
\item Demonstrates significant improvements in exploration time and robustness, when applied to state-of-the-art (SoTA) frameworks, as shown through extensive simulations; 
\item {Demonstrates real-world system resilience through a multi-trial dual-drone deployment, maintaining consistent exploration times despite communication, computation, mapping, and localization errors.}
\end{itemize}

The article's outline is as follows: Section \ref{sec:methodology} details the proposed enhancement to multi-robot exploration frameworks. The simulation experiments and analysis are presented in Section \ref{sec:results}, while the real world counterpart is presented in Section \ref{sec:real-world}. Finally, Section \ref{sec:conclusion} concludes the findings and proposes improvements to the proposed method.

\section{METHODOLOGY}

\label{sec:methodology}
\begin{figure}[h]
    \centering
    \includegraphics[width=0.5\textwidth]{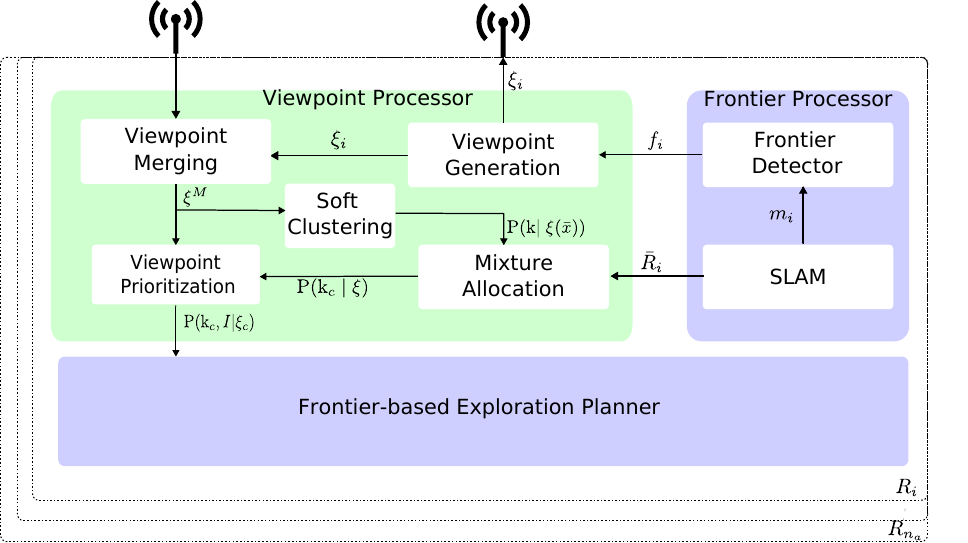}
    \caption{The proposed frontier prioritization module depicted as viewpoint processor processes the incoming frontiers and shares the output with a frontier-based exploration planner.}
    \label{fig:Methodology}
\end{figure}

An exploration team, $\mathcal{R}$, of $n_r$ robots is tasked to explore an unknown environment of area, $\mathcal{A}$. Each robot in $\mathcal{R}$ is individually referred to as $R_i$, for $i \in \{1,\dots ,n_r\}$, each with a pose $\bar{R_i}$. The robots are equipped with a perception sensor (e.g. LiDAR, depth sensors), with a perception range ($L$) and field of view ($\theta$), to generate a terrain representation. The proposed frontier prioritization solution is depicted as viewpoint processor in Fig.    \ref{fig:Methodology}. The frontiers generated by the frontier processor are shared with the viewpoint processor, which generates a probabilistic priority list that is shared with the exploration planner.\blfootnote{\href{https://youtu.be/d9iaTWqiM9c}{Video: https://youtu.be/d9iaTWqiM9c}}

\begin{figure*}[h]
    \centering
    \begin{subfigure}{0.32\textwidth}
        \centering
        \includegraphics[width=\textwidth]{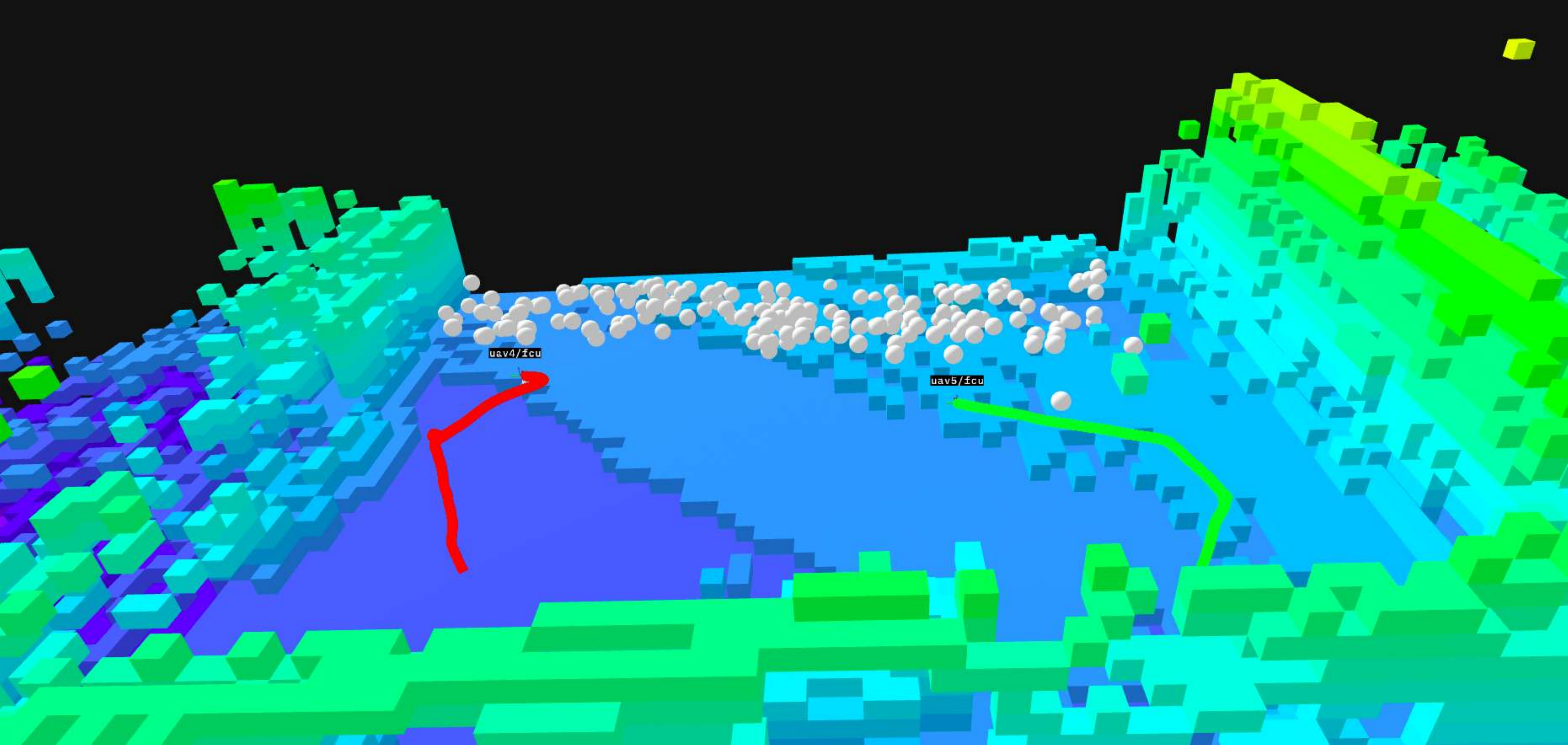}
        \caption{Merged Viewpoints}
        \label{fig:Real_World_Viewpoints}
    \end{subfigure}
    \hfill
    \begin{subfigure}{0.32\textwidth}
        \centering
        \includegraphics[width=\textwidth]{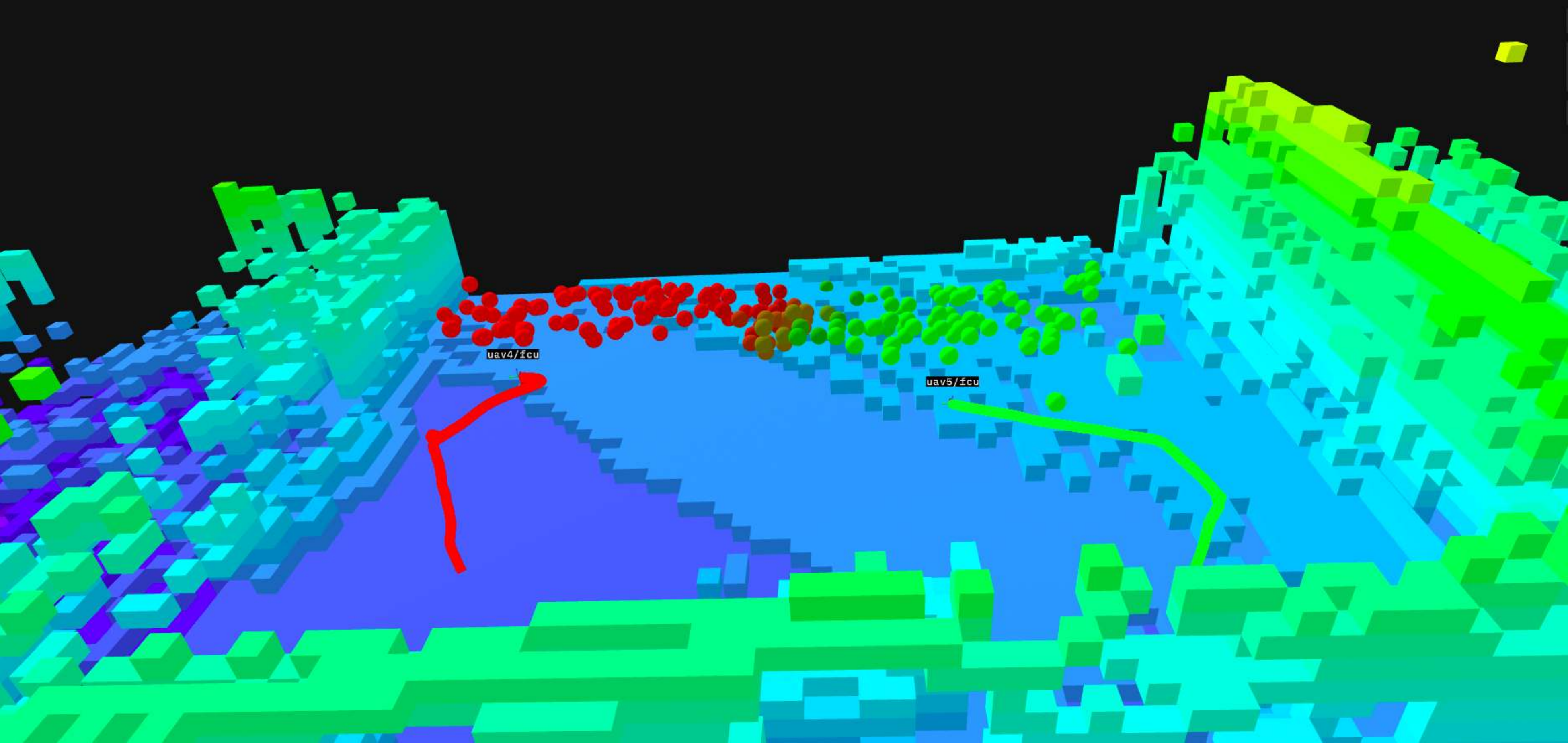}
        \caption{Soft Clustered Viewpoints}
        \label{fig:Real_World_GMMEM}
    \end{subfigure}
    \hfill
    \begin{subfigure}{0.32\textwidth}
        \centering
        \includegraphics[width=\textwidth]{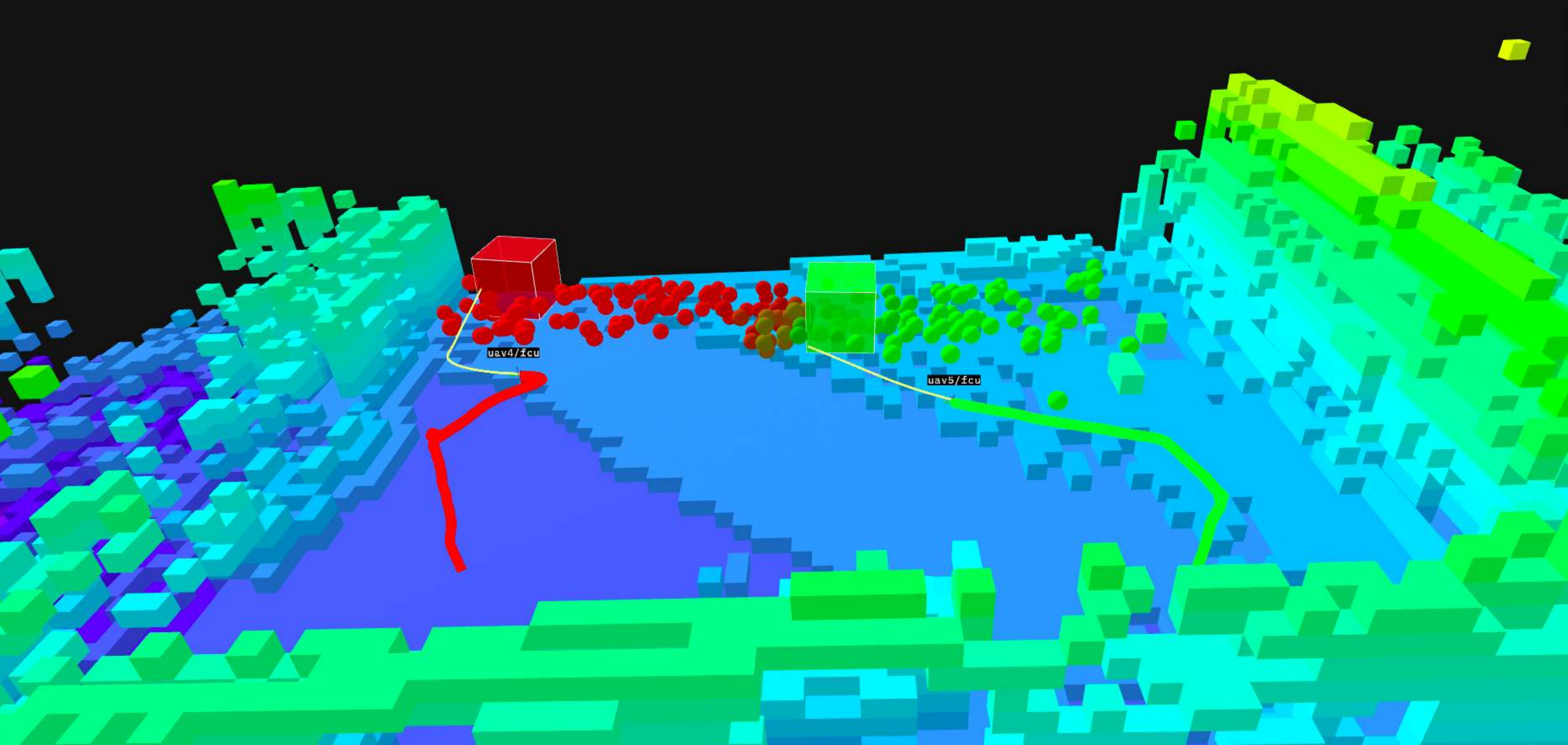}
        \caption{Prioritized Viewpoint}
        \label{fig:Real_World_Viewpoint_Prioritization}
    \end{subfigure}
    
\caption{An instance from a real-world experiment with $2$ UAVs in an Octomap. The UAV paths are shown in red and green. (a) The merged viewpoints, represented as white blobs, from Sec \ref{subsec:Viewpoint_Generation} (b) Soft cluster assignments from Sec \ref{subsec:Viewpoint_Clustering}, with color-coded probabilities visualized via smooth transitions between red and green to reflect each viewpoint's responsibility. (c) The prioritized viewpoint inferred from Sec \ref{subsec:Viewpoint_Priortization} is shown in red and green cubes.}
    \label{fig:Explainer}
\end{figure*}

\subsection{Frontier Processor}\label{subsec:Frontier_Processor}
The environment perceived by the perception sensor(s) of $R_i$ is used to build an ego-centric map, $m_i$,  generated by an onboard SLAM algorithm   \cite{palieri2020locus,xu2022fast}, which, if necessary, could also localize the agent within $m_i$, with pose $\bar{R_i}$. The frontier detector  \cite{keidar2014efficient,senarathne2013efficient} calculates the frontiers in $m_i$. The frontier processor provides algorithmic modularity, allowing the user to choose the ideal method depending on the robot type, the perception sensors, and the environment. A $2$D representation of frontiers is depicted in Fig.  \ref{fig:frontier_generation}.

\subsection{Viewpoint Processor}\label{subsec:Viewpoint_Processor}
\subsubsection{Viewpoint Generation and Merging}\label{subsec:Viewpoint_Generation}
Unlike previous works  \cite{lewis2024frontier,bartolomei2023fast}, where viewpoints are determined after clustering frontiers into groups and assigning a single viewpoint per cluster, our approach generates a dedicated viewpoint for each frontier. This fine-grained approach of generating a viewpoint per frontier, before any clustering, provides a richer set of candidate tasks, enabling a potentially optimal assignment compared to methods that assign a single viewpoint per cluster. Thus, at any given instant of exploration time, there is a set of viewpoints, $\xi_i$, associated with $R_i$. For the $j^{th}$ viewpoint in $\xi_i$, denoted as $\xi_{ij}$, is a tuple ($\bar{x}_{ij}$, $I_{ij}$), where $\bar{x}_{ij}$ denotes the  $\xi_{ij}$'s pose, and $I_{ij}$ represents the expected information gain generated from visiting  $\xi_{ij}$. For ease of notation, ($\bar{x}_{ij}$, $I_{ij}$) of each element in $\xi_i$ of $R_i$ will be represented as $(\xi_{i,j}(\bar{x}), \xi_{i,j}(I))$.

Assuming a binary classification of map cells into unknown ($0$) or known ($1$), the Shannon entropy of the map $m_i$ is given \cite{stachniss2005information} by
\begin{equation*} \label{eq:Entropy}
H(m_i) = - \sum_{c\in{m_i}}  p(c) \log_2 p(c) + (1-p(c)) \log_2(1-p(c)),
\end{equation*}
where $p(c)$ represents the probability that cell $c$ is occupied. 

{While a viewpoint is defined as a static pose from which frontiers can be covered, evaluating the true utility of visiting it requires considering the routed trajectory. Therefore, by computing the trajectory to each viewpoint using the A$^*$ algorithm, we can estimate the expected cumulative change in map cells from unknown to known along the path, up to and including the destination pose, considering the perception sensor's range and field of view.} Defining the final predicted map at the end of the trajectory to $\xi_{ij}$ as $\hat{m}_i^{\xi_{ij}}$, we can then compute the entropy $\hat{H}$ of $\hat{m}_i^{\xi_{ij}}$. The information gain, $I$, for visiting viewpoint, $\xi_{ij}$, can be estimated  \cite{stachniss2005information} as 


\begin{equation*} \label{eq:Information_Gain}
\xi_{ij}(I) = H(\hat{m}_i^{\xi_{ij}}) - H(m_i).
\end{equation*}

In   \cite{bartolomei2023fast}, the agents within the vicinity merge their maps to generate a global view. The choice of map/frontier merging is flexible depending on the underlying multi-robot exploration algorithm. With $\xi^M$ representing the merged viewpoints pooled in by $\mathcal{R}$, we can obtain a probability, $P(I|\xi)$, for each viewpoint $\xi \in \xi^M$, 
\begin{equation*} \label{eq:Information_gain_probability}
P(I \mid \xi) = \dfrac{\xi(I)}{\sum_{\xi \in \xi^M} \xi(I)} ,
\end{equation*}
 representing the probabilistic information gain for visiting $\xi$. An instance of the viewpoint merging is shown in Fig. \ref{fig:Real_World_Viewpoints}.

\subsubsection{Soft Clustering}\label{subsec:Viewpoint_Clustering}
To take a probabilistic approach to viewpoint prioritization, $\xi^M$ is clustered using a DP-GMM. Traditional GMM fits a convex combination of $K$ Gaussian distributions, each characterized by its mean and covariance, to a given dataset. In the context of viewpoint clustering, the probability, $P(k|\xi(\bar{x}))$, of a particular viewpoint, $\xi \in \xi^M$, belonging to the $k$-th Gaussian (with mean $\mu_k$ and covariance, $\Sigma_k$) is given by 

\begin{equation*} \label{eq:GMM}
P(k\mid\xi(\bar{x})) = \frac{\alpha_k \ \mathcal{N}(\xi(\bar{x}) \mid \mu_k, \Sigma_k)}{\sum_{j=1}^K \alpha_j \ \mathcal{N}(\xi(\bar{x}) \mid \mu_j, \Sigma_j)}, \quad \forall \xi \in \xi^M,
\end{equation*}
where $\alpha_k$ represents the weight (probability) of the $k$-th Gaussian, under the constraint $\sum_{j=1}^{K}\alpha_j = 1$ and estimated with expectation-maximization (EM)   \cite{moon1996expectation}. 

The number of viewpoints changes dynamically as the robot traverses through the map; fitting a predefined number of $K$ clusters with GMM can lead to erroneous distributions. While standard GMMs provide the desired soft clustering, they share the parametric constraint (fixed K) of rigid methods, such as $K$-means. We specifically utilize the Dirichlet process rather than the generalized Pitman-Yor process \cite{pitman1997two} to avoid unnecessary complexity, as the heavy-tailed cluster distributions characteristic of Pitman-Yor are not required for spatial frontier data. {Similar to GMM \cite{moon1996expectation}, DP-GMM fits the posterior distribution over the Gaussian parameters, the responsibility of each component, and determines the effective number of components. While various established approaches exist to solve this, such as Gibbs sampling  \cite{neal2000markov} or , we specifically employ variational inference \cite{blei2006variational} to efficiently determine this decomposition in our framework.} Detailed explanation of DP-GMM and the corresponding estimation algorithms would require more space than herein available; the interested reader is referred to  \cite{neal2000markov,blei2006variational}. Fig.\ref{fig:Real_World_GMMEM} showcases the soft clustering of the viewpoints in Fig.\ref{fig:Real_World_Viewpoints}. 

\subsubsection{Mixture Allocation}\label{subsec:Mixture_Allocation}
Fitting the robot pose, $\bar{R}$, to the DP-GMM model generated from viewpoint clustering yields a sequence of $k$ probabilities, each representing the responsibility of each component of the model for $\bar{R}$. The Gaussian component, $k_c$, that maximizes the probability represents the Gaussian component that is ideal for current $\bar{R_i}$ is given by  

\begin{equation*} \label{eq:chosen_cluster}
k_{c} = \argmax_{k} \alpha_k \mathcal{N}(\bar{R_i} \mid \mu_k, \Sigma_k).
\end{equation*}

\subsubsection{Viewpoint Prioritization}\label{subsec:Viewpoint_Priortization}
The Gaussian component $k_c$ shares a responsibility with all the viewpoints involved in the model-fitting. Thus, we can acquire the Gaussian probability distribution for the $k_c^{th}$ component by normalizing it:
\begin{equation*} \label{eq:GMM_chosen_Normalize}
P(k_c\mid\xi(\bar{x})) = \frac{ \mathcal{N}(\xi(\bar{x}) \mid \mu_{k_c}, \Sigma_{k_c})}{\sum_{\underline{\xi} \in \xi_i^M}  \mathcal{N}(\underline{\xi}(\bar{x}) \mid \mu_{k_c}, \Sigma_{k_c})}, \quad \forall \xi \in \xi_i^M.
\end{equation*}

{We define \textit{cluster coherence} as the responsibility of the Gaussian component, $k_c$, for a given viewpoint.} We can conclude that a high $P(k_c\mid\xi(\bar{x}))$  depicts a high cluster coherence to $k_c$.

The two probability distributions $P(I|\xi)$ and $P(k_c|\xi)$ represent two conditionally independent random variables seeded from a viewpoint, $\xi \in \xi_i^M$, namely, information gain and the cluster coherence. The joint probability 
\begin{equation*} \label{eq:JointProb}
P(k_c,I|\xi) = P(I|\xi) \times P(k_c|\xi)
\end{equation*}
informs $R_i$ of the potential gain of {the potential cumulative gain of navigating to and observing from viewpoint $\xi$}, while considering the ideal cluster, and can then be used in frontier-based multi-agent exploration algorithms \cite{bartolomei2023fast} to make an informed decision for trajectory generation.  The viewpoint with the highest priority amongst the viewpoints in Fig.\ref{fig:Real_World_Viewpoints} is portrayed in Fig.\ref{fig:Real_World_Viewpoint_Prioritization}. 

\subsection{Path planner}\label{subsec:Path_Processor}
To demonstrate the effectiveness and flexibility of our proposed enhancement, \textit{Frontier Prioritization (FP)}, we apply it to two distinct exploration frameworks: FAME \cite{bartolomei2023fast} and FroShe \cite{lewis2024frontier}. These frameworks were selected to represent disparate approaches: FAME employs cost optimization followed by an asymmetric traveling salesman problem (ATSP), whereas FroShe relies on bio-inspired heuristic swarm dynamics. Notably, FP is designed to be agnostic, making it compatible with any multi-robot frontier-based exploration algorithm.

\subsubsection{FAME}
FAME utilizes a dual-mode strategy, switching between an \textit{explorer} mode to push frontiers and a \textit{collector} mode to revisit \textit{islands}-small pockets of unknown space bypassed during initial exploration. This behavior is governed by an array of cost functions: coordination $J_{coord}$, distance $J_{D}$, velocity $J_{V}$, label $J_{L}$, and collaboration $J_{C}$. To integrate FP, we modify the collaboration cost, $J_{C}$. 

As detailed in \cite{bartolomei2023fast}, $J_{C}$ for a candidate viewpoint $\xi_{c}$, given peer UAV positions $\bar{R}_k$ and areas of interest $\mathcal{A}_{R}$, is defined as:
\begin{equation*}
 \kappa_{a}U_{a}(\xi_c, \tilde{x}_{R}^{i*}) + \kappa_{r}\sum_{k=0, k \ne i}^{N-1} \left(U_{r}(\xi_c, \tilde{x}_{R}^{k*}) + U_{r}(\xi_c, \bar{R}_k) \right),
\end{equation*}
where $(U_{a}, \kappa_{a})$ and $(U_{r}, \kappa_{r})$ represent the attractive and repulsive potential fields and their corresponding weights. These fields depend on the optimal assigned area $\tilde{x}_{R}^{k*}$ and current position $x_{R}^{k}$ of peer robot $k$. The first term represents the cost of attraction, while the summation accounts for repulsion. 

With FP, we replace the standard attraction term with $P(k_c, I | \xi_c)$ to directly incorporate the probabilistic reward of choosing the ideal candidate $\xi_c$. By introducing a weight $\kappa_{fp}$, the FP-enhanced collaboration cost $J^{fp}_{C}$ becomes
\begin{equation*}
    - \kappa_{fp} P(k_c, I | \xi_c) + \kappa_{r}\sum_{k=0, k \ne i}^{N-1} \left( U_{r}(\xi_c, \tilde{x}_{R}^{k*}) + U_{r}(\xi_c, x_{R}^{k}) \right).
\end{equation*}

\subsubsection{FroShe}
FroShe achieves robust multi-robot exploration by approximating frontiers as a virtual sheep swarm and applying shepherding heuristics\cite{lewis2024frontier}. The framework partitions the swarm into batches using hard K-Means clustering. These batches serve as the candidate targets, $\xi_c$, defined as tuples of the cluster centroid $\xi(\bar{x})$ and the cumulative cluster weight $\xi(w_b)$.

In the standard FroShe formulation, the optimal cluster to \textit{"herd"} is selected by minimizing travel distance while maximizing frontier weight:
\begin{equation*} 
\arg\max_{\xi \in \xi_c} \left( \lambda_m \frac{\xi(w_b)}{\xi(w_{max})} - \lambda_d \frac{\| \bar{R_i} - \xi(\bar{x}) \|_2}{d_{max}}\right) 
\end{equation*}
Here, the tunable weights $\lambda_m$ and $\lambda_d$ regulate the trade-off between information gain and traversal cost, normalized by the maximum candidate distance $d_{max}$ and weight $\xi(w_{max})$.
To integrate FP, we substitute the standard weight term with $P(k_c,I|\xi)$, utilizing its inherent encoding of information gain and cluster coherence. The resulting FP-based selection policy is
\begin{equation*} 
\arg\max_{\xi \in \xi_c} \left( \lambda_{fp} P(k_c, I | \xi_c)  - \lambda_d \frac{\| \bar{R_i} - \xi(\bar{x}) \|_2}{d_{max}}\right)
\end{equation*}
where $\lambda_{fp}$ is a tunable weight for prioritizing viewpoint cluster coherence and information gain over distance. {For our experiments, ($\lambda_{fp},\lambda_{m},\lambda_{d}$) are set as ($0.6,0.6,0.4$) to prioritize information gain over distance.}

\section{Simulation Analysis}
\begin{table*}
\centering
\renewcommand{\arraystretch}{1.3}
\resizebox{\textwidth}{!}{
\begin{tabular}{|c|c||c|c||c|c||c|c||c|c|}
\toprule
$n_r$ & Forest & \multicolumn{2}{c}{SPARSE FOREST (0.1 TREES / m$^2$)} & \multicolumn{2}{c}{MID-DENSITY FOREST (0.15 TREES / m$^2$)} & \multicolumn{2}{c}{DENSE FOREST (0.2 TREES / m$^2$)} & \multicolumn{2}{c}{MULTI-DENSITY FOREST} \\
\midrule
 & Variant & Standard & +FP & Standard & +FP & Standard & +FP & Standard & +FP \\
\midrule
\multirow[t]{3}{*}{1} & FAME & $724.63 \pm \underline{25.45}$ & $\textbf{687.56} \pm 38.47$ & $817.52 \pm \underline{26.89}$ & $\textbf{753.24} \pm 60.47$ & $829.10 \pm 70.29$ & $\textbf{776.13} \pm \underline{39.44}$ & $903.76 \pm \underline{18.30}$ & $\textbf{845.39} \pm 58.37$ \\
 & FroShe & $436.61 \pm 3.20$ & $\textbf{432.05} \pm \underline{2.88}$ & $462.57 \pm 3.59$ & $\textbf{454.71} \pm \underline{3.25}$ & $509.55 \pm 2.39$ & $\textbf{501.11} \pm \underline{2.18}$ & $544.54 \pm 2.45$ & $\textbf{536.72} \pm \underline{2.30}$ \\
 & Racer & $589.07 \pm 21.26$ & - & $648.46 \pm 19.60$ & - & $667.14 \pm 18.35$ & - & $761.59 \pm 69.18$ & - \\
\cline{1-10}
\multirow[t]{3}{*}{3} & FAME & $252.33 \pm 15.80$ & $\textbf{220.59} \pm \underline{13.54}$ & $274.13 \pm 13.60$ & $\textbf{251.84} \pm \underline{10.74}$ & $335.82 \pm 37.13$ & $\textbf{300.85} \pm \underline{12.40}$ & $389.63 \pm \underline{17.48}$ & $\textbf{345.67} \pm 25.52$ \\
 & FroShe & $154.79 \pm 2.41$ & $\textbf{142.21} \pm \underline{2.17}$ & $172.17 \pm 2.52$ & $\textbf{164.04} \pm \underline{2.23}$ & $226.01 \pm 2.96$ & $\textbf{224.98} \pm \underline{2.57}$ & $243.02 \pm 2.34$ & $\textbf{242.44} \pm \underline{2.12}$ \\
 & Racer & $401.09 \pm 50.27$ & - & $423.84 \pm 123.29$ & - & $509.68 \pm 66.36$ & - & $521.69 \pm 33.97$ & - \\
\cline{1-10}
\multirow[t]{3}{*}{4} & FAME & $183.17 \pm 13.58$ & $\textbf{160.74} \pm \underline{13.37}$ & $213.58 \pm \underline{12.27}$ & $\textbf{184.66} \pm 16.10$ & $261.27 \pm 22.14$ & $\textbf{230.92} \pm \underline{17.48}$ & $315.86 \pm 31.43$ & $\textbf{288.99} \pm \underline{16.54}$ \\
 & FroShe & $106.63 \pm 1.81$ & $\textbf{82.52} \pm \underline{1.73}$ & $121.34 \pm 1.64$ & $\textbf{111.24} \pm \underline{1.58}$ & $159.10 \pm 2.16$ & $\textbf{145.09} \pm \underline{1.98}$ & $193.88 \pm 2.20$ & $\textbf{191.97} \pm \underline{1.94}$ \\
 & Racer & $362.89 \pm 32.11$ & - & $395.95 \pm 46.54$ & - & $445.45 \pm 23.76$ & - & $448.08 \pm 98.96$ & - \\
\cline{1-10}
\multirow[t]{3}{*}{6} & FAME & $118.18 \pm 12.69$ & $\textbf{95.11} \pm \underline{4.12}$ & $145.48 \pm 10.53$ & $\textbf{121.01} \pm \underline{9.06}$ & $178.72 \pm 33.83$ & $\textbf{150.44} \pm \underline{10.48}$ & $217.00 \pm 20.56$ & $\textbf{194.20} \pm \underline{12.69}$ \\
 & FroShe & $75.99 \pm \underline{1.60}$ & $\textbf{56.47} \pm 1.65$ & $93.55 \pm 1.41$ & $\textbf{65.12} \pm \underline{1.28}$ & $114.92 \pm 2.02$ & $\textbf{101.91} \pm \underline{1.82}$ & $139.53 \pm 2.05$ & $\textbf{121.19} \pm \underline{1.74}$ \\
 & Racer & $225.64 \pm 41.49$ & - & $224.26 \pm 99.28$ & - & $285.48 \pm 49.10$ & - & $264.94 \pm 17.79$ & - \\
\cline{1-10}
\multirow[t]{3}{*}{8} & FAME & $85.31 \pm 13.11$ & $\textbf{66.61} \pm \underline{5.05}$ & $104.53 \pm 16.46$ & $\textbf{82.68} \pm \underline{5.82}$ & $140.68 \pm 12.62$ & $\textbf{117.85} \pm \underline{5.81}$ & $166.35 \pm \underline{11.44}$ & $\textbf{142.87} \pm 12.11$ \\
 & FroShe & $63.03 \pm 1.51$ & $\textbf{51.12} \pm \underline{1.36}$ & $82.23 \pm 2.20$ & $\textbf{64.65} \pm \underline{1.78}$ & $113.94 \pm 1.69$ & $\textbf{97.70} \pm \underline{1.62}$ & $122.91 \pm 1.53$ & $\textbf{105.33} \pm \underline{1.34}$ \\
 & Racer & $130.42 \pm 23.85$ & - & $180.88 \pm 46.02$ & - & $223.91 \pm 21.32$ & - & $252.29 \pm 18.55$ & - \\
\cline{1-10}
\multirow[t]{3}{*}{10} & FAME & $70.16 \pm 9.52$ & $\textbf{50.64} \pm \underline{4.40}$ & $88.38 \pm 16.25$ & $\textbf{65.71} \pm \underline{4.97}$ & $114.81 \pm 16.42$ & $\textbf{101.33} \pm \underline{6.98}$ & $132.24 \pm 12.28$ & $\textbf{105.66} \pm \underline{6.88}$ \\
 & FroShe & $\textbf{51.18} \pm 1.27$ & $53.36 \pm \underline{1.16}$ & $64.47 \pm 2.17$ & $\textbf{53.20} \pm \underline{1.98}$ & $83.74 \pm 1.59$ & $\textbf{79.00} \pm \underline{1.48}$ & $104.46 \pm 1.64$ & $\textbf{87.73} \pm \underline{1.42}$ \\
 & Racer & $100.84 \pm 19.11$ & - & $110.02 \pm 9.69$ & - & $200.49 \pm 19.14$ & - & $190.96 \pm 10.04$ & - \\
\cline{1-10}
\bottomrule
\end{tabular}}
\caption{Mean successful exploration times for FAME, RACER, and FroShe across varying UAV team sizes ($n_r$) and forest densities. The Standard and +FP columns represent the default and FP-enabled frameworks, respectively. The lowest mean value is \textbf{bolded}, and the lowest standard deviation is \underline{underlined}.}
\label{tab:simulation_mixed_density}
\end{table*}
\label{sec:results}
\subsection{Setup}
To ensure a consistent analysis, we utilize four simulated forest environments \cite{bartolomei2023fast} of size $50$m $\times$ $50$m $\times$ $2$m  of sparse ($0.1$trees/$m^2$), mid ($0.15$trees/$m^2$), high ($0.2$trees/$m^2$), and mixed density are used for our analysis, similar to  \cite{bartolomei2023fast}. All environments are tested with ($1, 3, 4, 6, 8, 10$) drones across $10$ iterations. To ensure fair comparisons that respect the original design of each algorithm, perception sensors differ by framework: FAME, FAME+FP, and RACER utilize a $640\times480$ pixel depth camera with an $80^\circ\times60^\circ$ field of view, whereas FroShe and FroShe+FP employ a $360^\circ$ FoV $3$D-LiDAR. This sensor variety also strengthens the case for FP's robustness to sensor choice. The choice to include RACER serves to demonstrate that FAME+FP improves exploration time even in scenarios where the baseline FAME typically underperforms against RACER. Scikit-learn's \cite{scikit-learn} DP-GMM module is used for our soft frontier clustering, with the maximum number of clusters bounded to half the frontier size. We enforce the covariance matrix to be diagonal to ensure faster computation and minimize over-fitting. 

\begin{figure*}[h]
    \centering
    \begin{subfigure}[b]{0.32\textwidth}
        \centering
        \includegraphics[width=\textwidth]{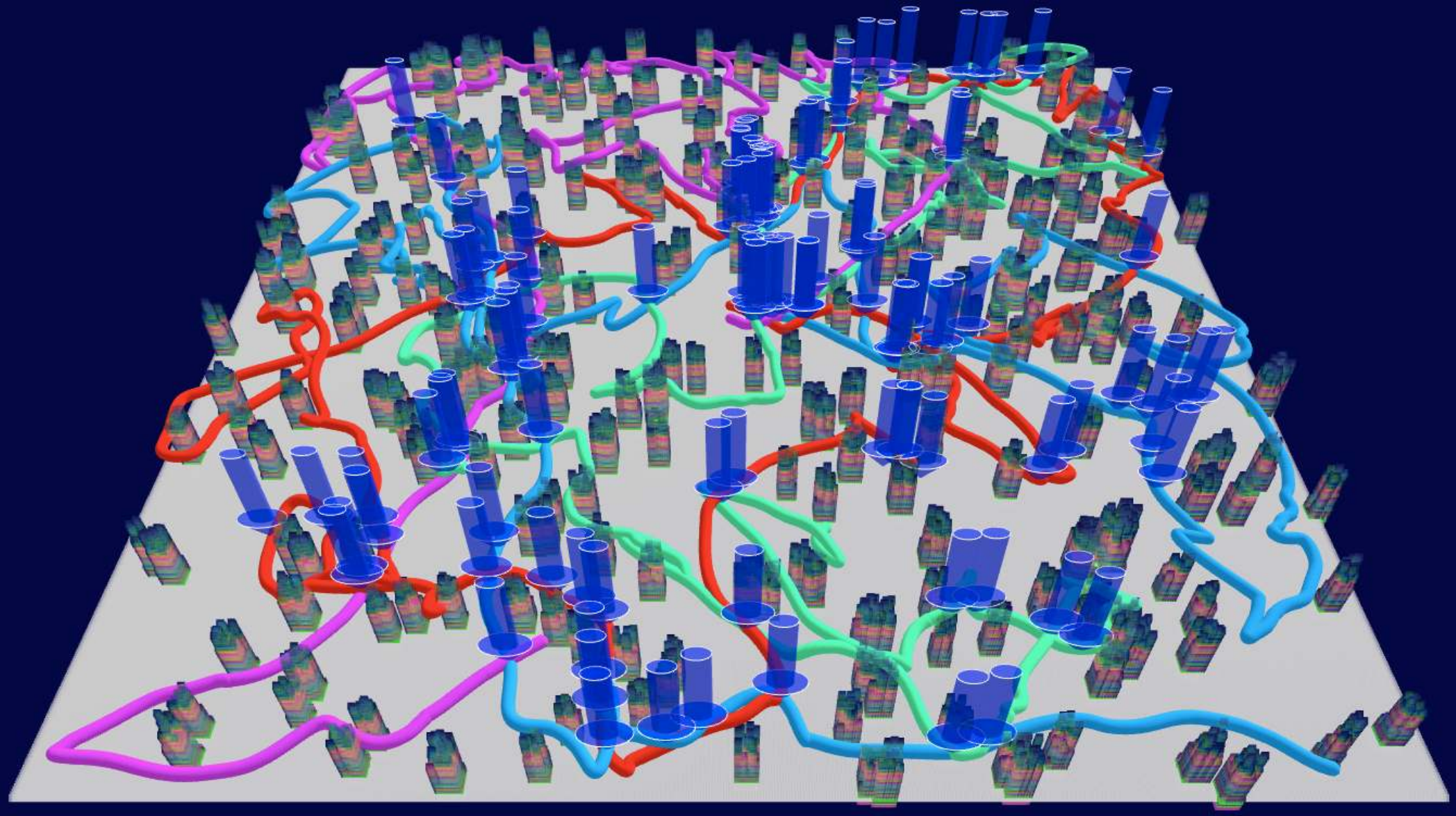}
        \caption{{FAME UAV Trajectories}}
        \label{fig:FAMEOverlaps}
    \end{subfigure}
    \hfill
    \begin{subfigure}[b]{0.32\textwidth}
        \centering
        \includegraphics[width=\textwidth]{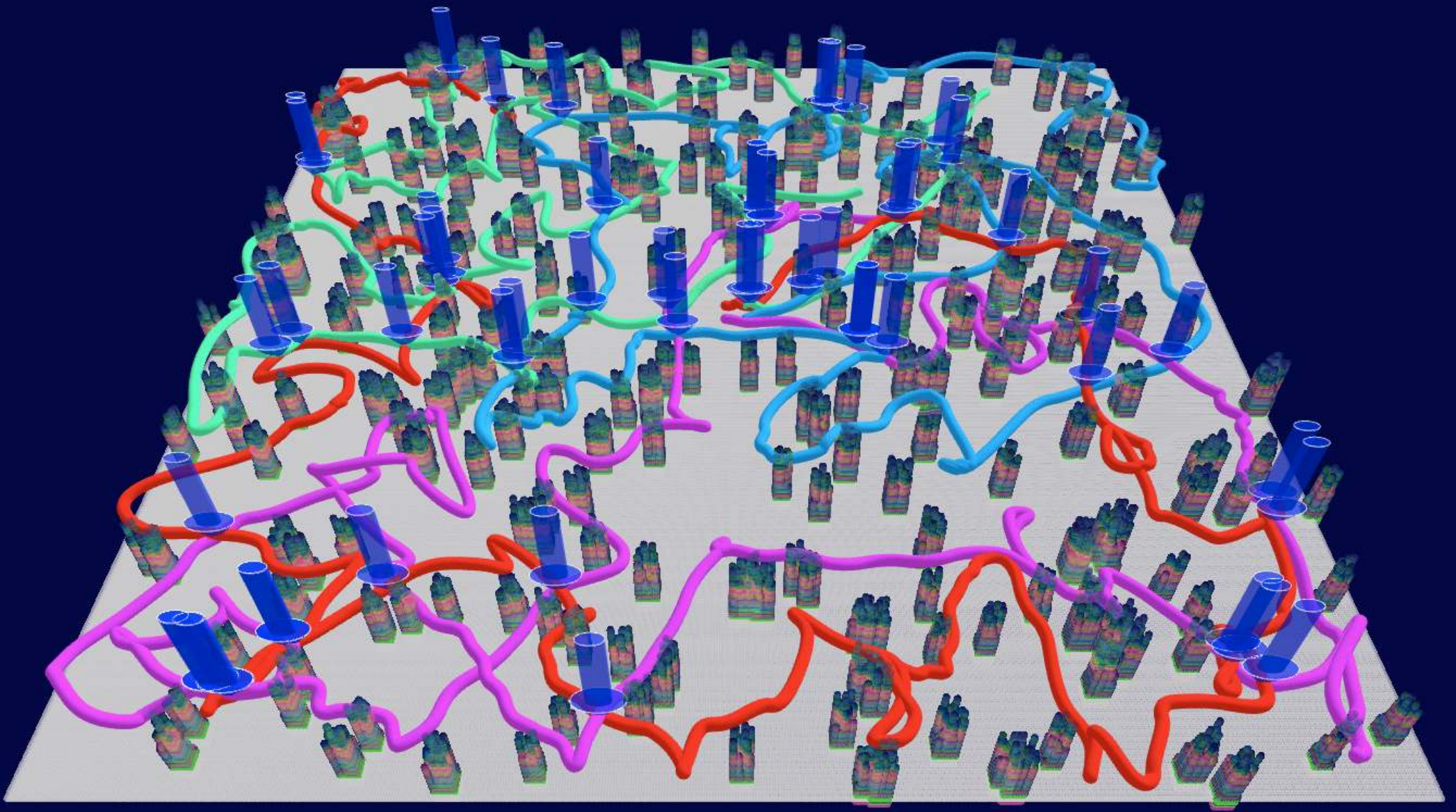}
        \caption{{FAME+FP UAV Trajectories}}
        \label{fig:FAMEFPOverlaps}
    \end{subfigure}
    \hfill
\begin{subfigure}[b]{0.32\textwidth}
    \centering
    \includegraphics[width=\textwidth,height=0.6\textwidth]{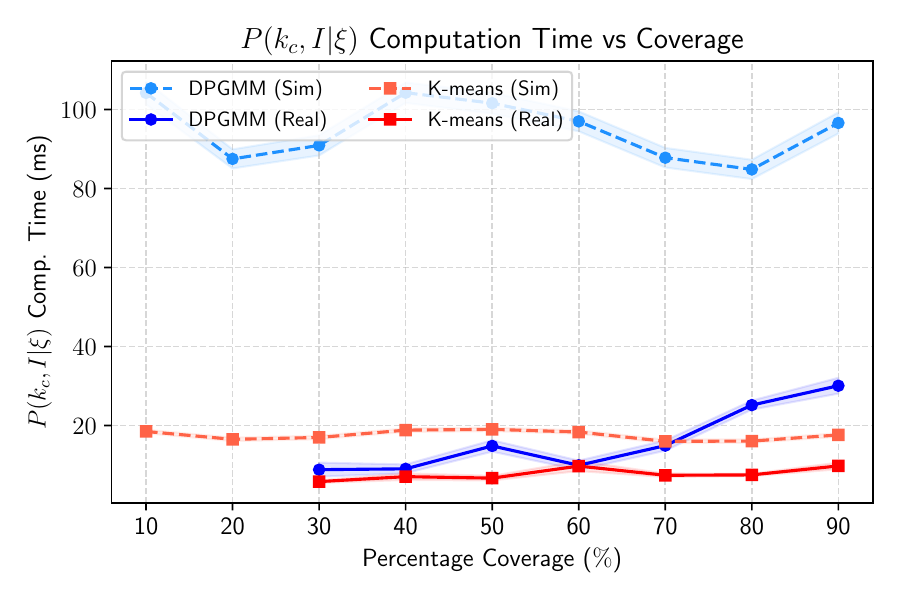}
    \caption{{Comp. Time vs Exploration Coverage.}}
    \label{fig:ComputationLoad}
\end{subfigure}
    
\caption{(a) and (b) showcase the trajectories completed by $4$ UAVs in a $50m \times 50m$ mixed density forest simulation using FAME and FAME+FP respectively. The blue arrows showcase the various points at which UAV trajectories intersect.\\ (c) Showcases the computational time required to calculate $P(k_c,I|\xi)$ across the mission coverage for a dual-drone system.}
\label{fig:OverlapsandComputation}
\end{figure*}

\subsection{Varying Obstacle Clutter}
Tab. \ref{tab:simulation_mixed_density} consolidates the exploration time analysis for varying fleet sizes ($n_r$) and forest densities under unrestricted communication. The inclusion of FP showcases a consistent improvement across both FAME and FroShe, averaging approximately $14\%$ and $10\%$ respectively. 

In the single-agent setting, FAME alone falls short of RACER's efficiency. However, the integration of FP significantly mitigates this disparity, narrowing the performance gap to RACER from ($19\%,20\%,23\%,16\%$) to ($11\%,12\%,15\%,6\%$) for the sparse, mid, high, and mixed density forests. This suggests that FP effectively prioritizes viewpoint thanks to an improved $J_c$ formulation with $P(k_c,I|\xi)$.

In multi-UAV experiments, with enhanced viewpoint prioritization, FAME+FP demonstrates robust scalability, yielding a minimum time reduction of $10\%$ with $3$ drones in dense forests, and peaking at a significant $23\%$ improvement with $10$ drones in sparse environments. Similarly, while FroShe already achieves fast exploration via its $360^\circ$ sensor, adding FP yields further gains, reaching a $25\%$ improvement for $6$ UAVs. Notably, these benefits begin to diminish for FroShe with $8$ to $10$ UAVs due to sensing redundancy. While higher environment size would likely alleviate this saturation, such large-scale validation is reserved for future work.

Fig. \ref{fig:FAMEOverlaps} illustrates the trajectory overlaps (when one UAV crosses the traversed path of its peer) for a single simulation instance. The standard FAME approach results in $108$ overlaps, whereas the addition of Frontier Prioritization (Fig. \ref{fig:FAMEFPOverlaps}) lowers this count to $54$. Consequently, the simulation data shows a reduction in total distance traveled: FAME+FP achieved a mean path length of $270.34$m, compared to $312.39$m for the baseline.
\begin{figure*}[ht]
    \centering
    \includegraphics[width=\textwidth,height=0.25\textwidth]{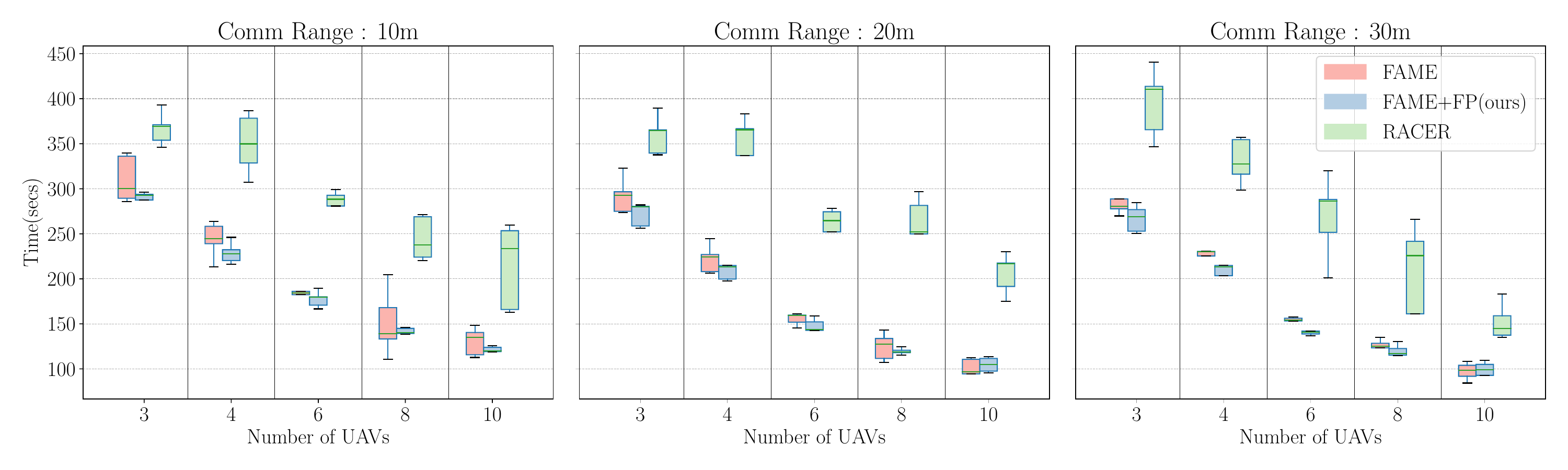}
    \caption{Exploration time analysis across a dense forest environment with varying communication range and UAV count.}
    \label{fig:comm_time_analysis}
\end{figure*}

{To isolate the performance gains of soft clustering (Sec. \ref{subsec:Viewpoint_Clustering}) from the advantage of evaluating a larger number of viewpoints, we implement high-density hard-clustering K-means module to replace the soft clustering module. To ensure a fair comparison, the number of clusters K is fixed to the maximum number of clusters observed during the corresponding DP-GMM trials. To enforce hard clustering binary assignments, the probabilistic cluster coherence $P(k_c|\xi)$ is $1$ if the viewpoint $\xi$ belongs to the cluster, and $0$ otherwise. Analysis is carried out in the simulated forest environments, as well as in a simulation utilizing the SLAM map generated from the real-world basketball court experiment (detailed in Section \ref{sec:real-world}). Averaged across $10$ runs for a dual drone system, the DP-GMM framework demonstrated significant reductions in exploration time compared to the high-density K-means baseline: ($23.3\%,22.5\%,32.0\%,23.5\%,30.3\%$) in the (basketball court, sparse forest, mid-density forest, dense forest, multi-density forest).}

{The operational advantage of DP-GMM is most evident during the late stages of exploration. At this phase, the number of active frontiers drops significantly. Due to its strict delegation, hard clustering fails here, it prioritizes only a small subset of frontiers for each robot, while the other frontiers are disregarded. As the frontier count dwindles, these rigid clusters force multiple UAVs to converge on the same remaining viewpoints. With such a limited pool of choices, the robots inevitably select overlapping trajectories that degrade FAME's performance. In contrast, DP-GMM ensures all frontiers remain part of the prioritization process. Because the candidate set is never "subsidized" or artificially restricted, the team prioritizes different frontiers. By preventing this bottleneck, DP-GMM-based prioritization improves FAME where K-Means would otherwise cause it to fail.}


\subsection{Computation Load Analysis}
To address computational constraints, we implement a multi-tiered optimization strategy. First, we enforce diagonal covariances during the EM process. This reduces the time complexity of fitting $K$ components to $N$ data points from $\mathcal{O}(TNKd^2)$ to $\mathcal{O}(TNKd)$, where $d$ is the dimensionality and $T$ is the convergence iterations. Second, to prevent the clustering latency from affecting the robot's real-time performance, the prioritization module operates as a standalone, asynchronous ROS node. This decouples the exploration logic from the critical control loop. We further reduce the computational load by employing warm starts for the DP-GMM, minimizing the iterations required for convergence. 

{Furthermore, to evaluate real-time operational feasibility, Fig. \ref{fig:ComputationLoad} compares the computational latency of DP-GMM against high-density K-Means in both a real-world basketball environment and its digital twin. The results indicate that, aided by the proposed optimizations, DP-GMM exhibits a consistent computational cost throughout the mission. Additionally, for a given hardware setup and real-world dataset, DP-GMM and K-Means demonstrate comparable execution times. To analyze scalability regarding frontier count, we increased the frontier resolution in the simulation. Notably, an approximate $20\times$ increase in the number of frontiers resulted in only a $5\times$ relative increase in DP-GMM's computational overhead compared to K-Means. To overcome the interpreter overhead of the Python-based scikit-learn implementation, future work will leverage C++ alternatives like BayesMix \cite{beraha2025bayesmix} to synchronize execution speeds with SLAM updates.}

\begin{figure*}[h]
    \centering
    \begin{subfigure}{0.32\textwidth}
        \centering
        \includegraphics[width=\textwidth,height=0.5\textwidth]{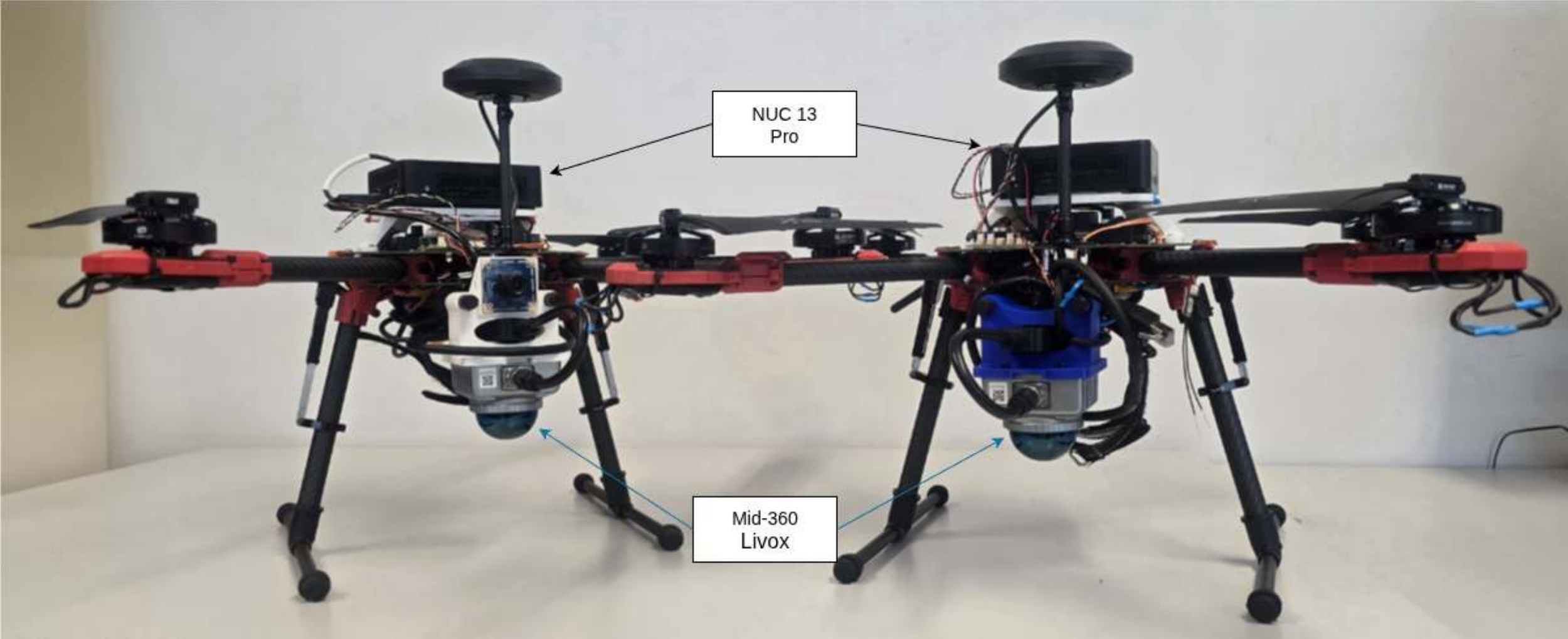}
        \caption{Two Tarot $650$ Sport}
        \label{fig:Real_World_Drones}
    \end{subfigure}
    \hfill
    \begin{subfigure}{0.32\textwidth}
        \centering
        \includegraphics[width=\textwidth,height=0.5\textwidth]{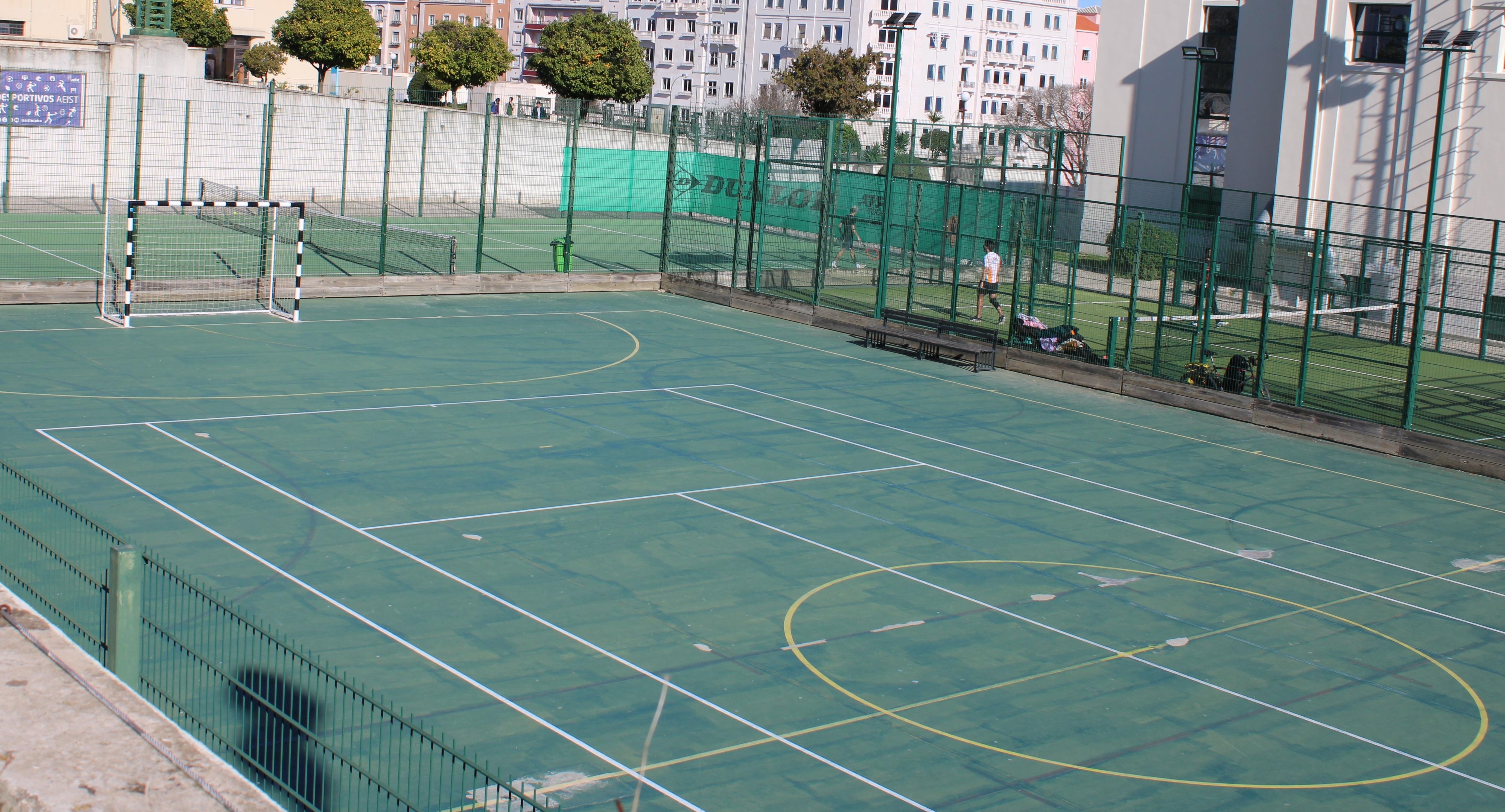}
        \caption{Test Environment}
        \label{fig:Real_World_Test_Environment}
    \end{subfigure}
    \hfill
    \begin{subfigure}{0.32\textwidth}
        \centering
        \includegraphics[width=\textwidth,height=0.5\textwidth]{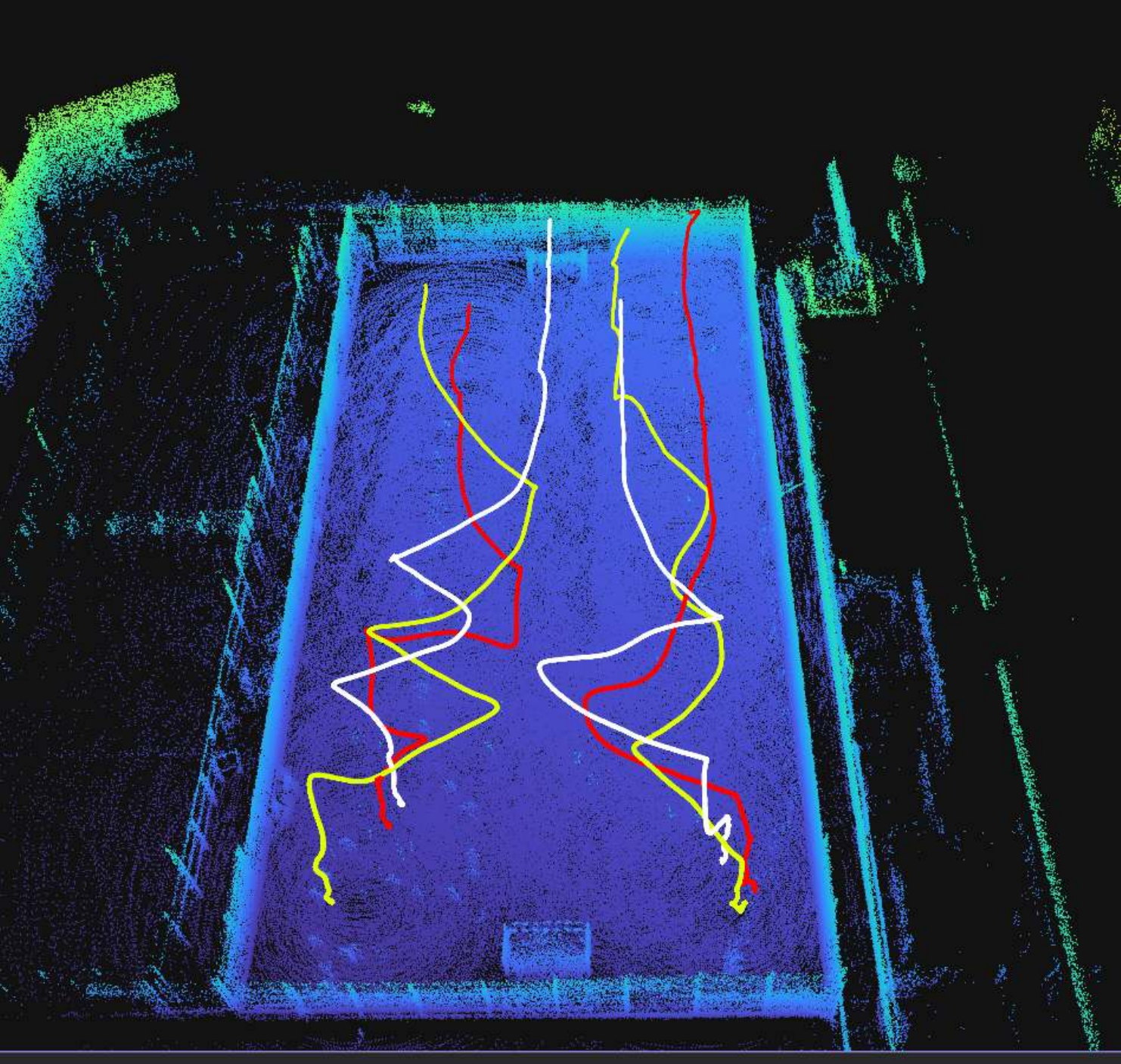}
        \caption{Pointcloud Reconstruction}
        \label{fig:Real_World_PCL}
    \end{subfigure}    
\caption{(a) Two Tarot 650 UAVs, each equipped with Mid-360 Livox and ASUS NUC Pro. (b) A minimal feature $18m\times36m\times3m$ volume test environment. (c) The reconstructed point cloud and the path taken across $3$ trials.} 
    \label{fig:Real_World}
\end{figure*}

\subsection{Communication Constrained Analysis}
The box plots in Fig. \ref{fig:comm_time_analysis} depict the exploration time analysis for FAME, FAME+FP, and RACER for the dense forest, with communication thresholds of $10$m, $20$m, and $30$m. Consistent with the enhancement seen in Tab. \ref{tab:simulation_mixed_density}, the enhancement improves FAME's exploration across varying team sizes. Furthermore, FAME exhibits high variance in exploration time in scenarios with communication constraints of $10$m and $20$m. With the proposed enhancement, we see a reduction in the standard deviation owing to better delegation of clusters.



\section{Real-World Experiments}

\label{sec:real-world}
To demonstrate end-to-end feasibility and real-time operation, the proposed frontier prioritization method was implemented on a dual-drone setup to explore an $18m\times36m\times3m$ volume as shown in Fig. \ref{fig:Real_World_Test_Environment}. {While simulation benchmarks provided head-to-head performance comparisons, the primary objective of the real-world experiments was to validate the robustness of the frontier prioritization under physical uncertainties. Factors such as LiDAR noise, drift in state estimation, imperfect map fusion, and intermittent communication introduce constraints that are minimal in simulation, but often disruptive in the real-world.}

Each $T650$-sport UAV is equipped with Mid-$360$ Livox lidar for perception and an ASUS NUC $13$ Pro for processing, as seen in Fig. \ref{fig:Real_World_Drones}. The whole framework is deployed on an Apptainer running ROS Noetic on Intel Core i$7 - 1360$P CPU @ $2.2$GHz$\times 16$. The underlying FAME+FP algorithm is configured with a maximum horizontal [velocity, acceleration] of each UAV is limited to [$1m/s$, $1m/s^2$]. Each UAV runs an onboard SLAM~\cite{palieri2020locus} to generate the 3D representation of the environment. The LiDAR range is limited to $10m$ in the x-direction, while it is unrestricted in the y-direction. An unrestricted y is chosen to ensure safety in the confined test environment, while x is constrained to mandate collaborative motion to explore the complete volume. The UAVs exchange their SLAM maps over WiFi to construct a shared global 3D OctoMap~\cite{hornung2013octomap}, from which frontiers are extracted for path planning. The MRS-UAV framework~\cite{baca2021mrs} is used for trajectory generation~\cite{kratky2021autonomous}, obstacle avoidance, and safety-critical inter-UAV collision avoidance~\cite{baca2018model}. Instances of the pipeline are presented in Fig. \ref{fig:Explainer}.

A trial is considered successful if $95\%$ of the volume is explored by the UAVs. Across the three trials, exploration to $95\%$ coverage was achieved in $(77,83,85)$ seconds. 
Despite the aforementioned real-world uncertainties, the path analysis in Fig. \ref{fig:Real_World_PCL} reveals a distinct absence of overlapping paths. This result indicates that the method produces a stable, non-redundant spatial division of labor without requiring explicit trajectory negotiation or coupling. The UAVs effectively adhered to their preferred clusters even when subjected to real-world sensing and localization errors. This confirms that the joint-probabilistic frontier prioritization model successfully balances information gain with cluster coherence, ensuring efficient coordination in practical deployments where communication is limited and negotiation is costly.


\section{CONCLUSIONS AND FUTURE WORK}

\label{sec:conclusion}
This article proposes an enhancement to improve frontier-based multi-robot exploration frameworks. The enhancement uses the non-parametric Dirichlet process Gaussian mixture model to achieve frontier clustering and probabilistic information gain for smooth exploration task delegation. The proposed enhancement is tested on existing SoTA frontier-based multi-robot exploration algorithms for forests, and the results showcase an improvement in overall time efficiency and consistency. The simulation analysis was carried out in varying environment clutter, communication constraints, and team sizes, with the enhanced versions consistently outperforming the original method. In sparse forests, we observed up to a $23\%$ reduction in exploration time with a $10$ UAV system for FAME \cite{bartolomei2023fast}, and $25\%$ improvement for a $6$ UAV system for FroShe \cite{lewis2024frontier}. {Complementing the algorithmic benchmarks established in the simulated forests, deployment of the proposed method in a dual-drone system validated the framework's end-to-end system integration. Real-world trials successfully showcased consistent exploration times despite errors arising from computational delays, communication delays, incorrect map merging, and erroneous pose estimates; confirming that the soft probabilistic clustering maintains stable task delegation even under real-world uncertainties.}
A future extension of the work can aim to minimize the CPU load by a more efficient Bayesian library \cite{beraha2025bayesmix} for determining the mixtures.


%





\printbibliography

\end{document}